\documentclass[letterpaper, 10 pt, conference]{ieeeconf} 

\usepackage{lmodern}
\usepackage{graphicx}
\usepackage{times}
\usepackage{ragged2e}
\usepackage{amsmath,amssymb}
\usepackage{tabularx}
\usepackage{hyperref}
\usepackage[ruled,vlined]{algorithm2e}
\usepackage{subfig}
\usepackage{dblfloatfix}
\usepackage{authblk}

\usepackage{xcolor}

\usepackage{soul}
 
\begin{document}
\noindent

\title{\LARGE \bf
Learning-based Observer Evaluated on the Kinematic Bicycle Model
}

\author{Agapius Bou Ghosn$^{1}$, Philip Polack$^{1}$ and Arnaud de La Fortelle$^{1}$
\thanks{$^{1}$Agapius Bou Ghosn, Philip Polack and Arnaud de La Fortelle are with Center for Robotics, MINES ParisTech, PSL University, 60 Bd. Saint Michel 75006 Paris, France}}

\maketitle

\thispagestyle{empty}
\pagestyle{empty}

\begin{abstract}
The knowledge of the states of a vehicle is a necessity to perform proper planning and control. These quantities are usually accessible through measurements. Control theory brings extremely useful methods -- observers -- to deal with quantities that cannot be directly measured or with noisy measurements. Classical observers are mathematically derived from models. In spite of their success, such as the Kalman filter, they show their limits when systems display high non-linearities, modeling errors, high uncertainties or difficult interactions with the environment (e.g. road contact). In this work, we present a method to build a learning-based observer able to outperform classical observing methods. We compare several neural network architectures and define the data generation procedure used to train them. The method is evaluated on a kinematic bicycle model which allows to easily generate data for training and testing. This model is also used in an Extended Kalman Filter (EKF) for comparison of the learning-based observer with a state of the art model-based observer. The results prove the interest of our approach and pave the way for future improvements of the technique.
\end{abstract}

\section{Introduction}
In control applications, a difference of behavior exists between the actual system and its model. This difference of behavior is caused by the assumptions underlying the model and by external disturbances known as process noise. Similarly, a difference exists between the measurements of the actual system and its actual state and is caused by measurement noise. To tackle this problem, state observation is a necessity to estimate the internal state of a system. State estimation is of great importance to the motion planner and controller, and the estimation quality has direct influence on the system's control quality, knowing that the estimations are used in the control process.

Classical observing techniques are used for state estimation purposes. They rely on a model of the system in order to predict the future states based on the current ones, the inputs and the provided measurements. The used system model should be as close as possible to the actual system behavior to produce accurate estimations.

Classical observers are widely used in the literature in many control applications; we present next the different types of classical observers applied in the literature to vehicle state estimation. The Luenberger observer, which is a linear observer that uses the linear state space representation of a system to predict its state with a correction term that includes the difference between the predicted observations and the measurements, has been employed in applications like \cite{cherouat_vehicle_2005}, \cite{kiencke_observation_1997}, to estimate the vehicle velocity, side slip angle, and yaw rate using a dynamic bicycle model and a linear tire model. Limitations occur when the vehicle steps out of the linear operational domain. To overcome nonlinearity issues, nonlinear observers are used. They involve the update equations of the model in addition to a correction term to estimate a system's state. Applications included estimation of the side slip angle as in \cite{noauthor_estimation_2022}, \cite{lee_slip_2013}, or the vehicle's velocity as in \cite{zhao_design_2011}, \cite{imsland_vehicle_2006}; In these works, the vehicle is described by either a dynamic bicycle model or a four wheel dynamic model associated with different tire models; the model used is then more representative of a vehicle's maneuvers, providing more accurate estimations in nonlinear cases.

Other applications use the Kalman filter for state observation. The Kalman filter is model based and deals with linear systems. Its algorithm is split into two steps: the prediction step where it predicts the current state and uncertainty based on the previous state, applied inputs and covariances and the update step where it corrects the predicted state based on the current measurements and covariances. The Extended Kalman Filter is an extension to the Kalman filter to deal with nonlinear systems. It performs linearizations at each step around the current estimate. The EKF has been used along with a dynamic bicycle model or a four wheel dynamic model to estimate the vehicle's velocity, yaw rate and tire forces as in \cite{kim_vehicle_2018}, \cite{wilkin_use_2006}, or for slip angle estimation as in \cite{sun_research_2022}, \cite{reina_vehicle_2019} with different considerations to the choice of the tire model.

As presented so far, observers are able to estimate multiple variables of interest based on models and measurements. Presented techniques showed accurate results in a specific operational range, the problem lies in their validity beyond the assumptions of the model they use. Classical observers use vehicle and tire models to represent the motion of the vehicle: these are subject to their own assumptions. In other words, model based observers are limited to the validity of the model (vehicle and tire models) used in their estimations; any change in the hypotheses that define the model would result in the degradation of the observer's performance.

In several works, learning methods have integrated in observers, either in parallel with vehicle models as in \cite{liu_vehicle_2021} or alone as in \cite{chindamo_estimation_2018}. The methods used resulted in accurate estimations but either were still depending on a vehicle model, failed to converge in many scenarios, or weren't compared to a known benchmark. 

In this paper, we present a complete framework to create a robust learning based observer, including the neural network architecture and the data generation methods. The observer requires a lot of training data of the system (inputs and ground truth); and also testing data. In this preliminary work, we intend to show the condition of feasibility of our method.

The observer gets as input what is known: measures and control inputs. The output is what we would like to observe: it could be any physical quantity linked to the system and even other features (like uncertainties or context related indicators). Since there is no model, the observer is not limited to estimation of the state of the model (or parameters). The difficulty is to get a ground truth for these quantities.

Therefore we show in this paper a simple case, where the system is modeled as a kinematic bicycle, a simple non-linear model. This allows to generate data by simulation (including the ground truth) and also to compare with a state of the art observer, the Extended Kalman Filter (EKF).

The rest of the paper is organized as follows: Section \ref{kbm.sec} presents the kinematic bicycle model which will be used as a simulator for our system. This model is also used for the prediction step of the reference EKF observer.
Section \ref{3.sec} presents the learning based observer including data generation algorithms. Section \ref{4.sec} compares the performance with the EKF with a detailed error analysis. {Section~\ref{5.sec}} concludes the paper.

\section{The kinematic bicycle model}\label{kbm.sec}
As it was stated in the introduction, a learned observer does not necessarily require a model, but one is used in this paper for two purposes: on the one hand, simulation for generating datasets for training and testing; on the other hand for the prediction step of the reference EKF observer used for comparison.

\begin{figure}[ht]
\begin{center}
\includegraphics[width=.8\columnwidth]{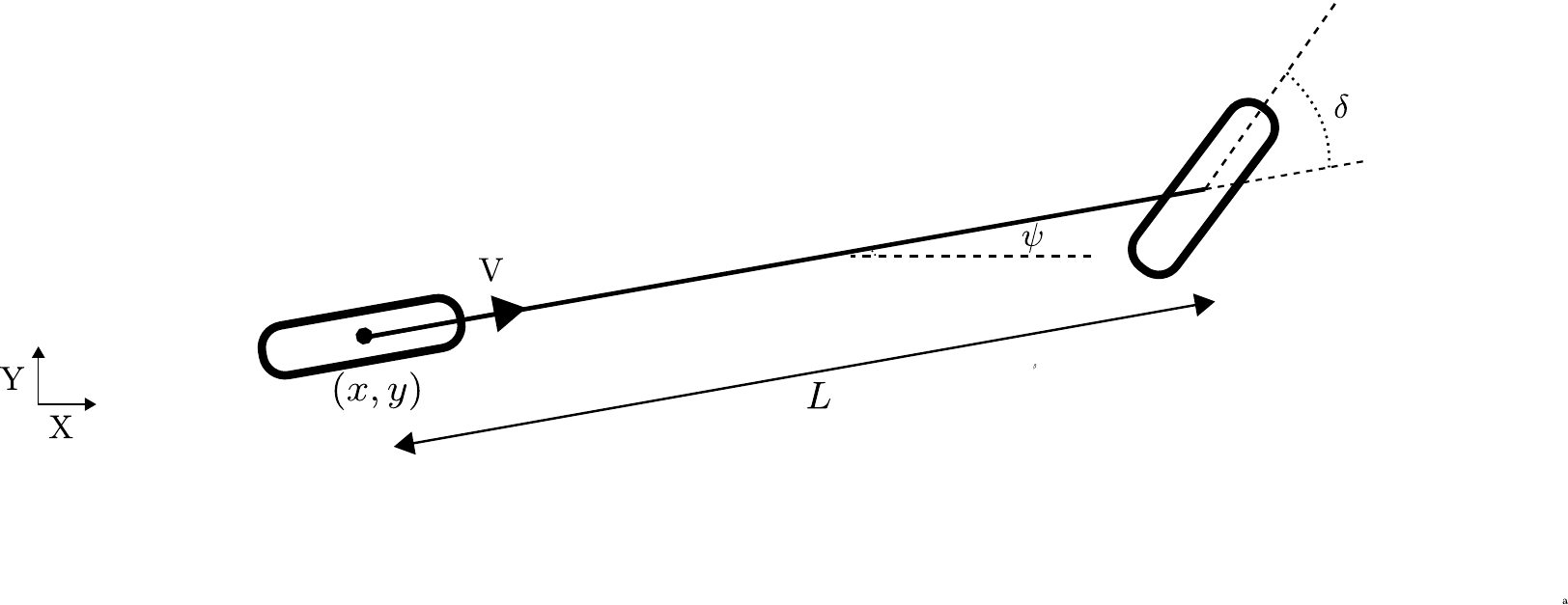}
\caption{The kinematic bicycle model. }
\label{kinematicModel.fig}
\end{center}
\end{figure}

In the rest of the paper, the system is identified to the kinematic bicycle model, described in Figure \ref{kinematicModel.fig}. The kinematic bicycle model is used in many vehicle planning and controlling applications in the literature (e.g. \cite{polack_guaranteeing_2018}). This model is a simplified one that does not consider dynamics such as forces, masses and inertia. Several assumptions define this model \cite{rajamani_lateral_2012}: first, the wheels of the vehicle are lumped into a bicycle model; second, slope and road bank angles are neglected; third, the pitch, roll and vertical dynamics are neglected; fourth, the model is only valid for low speed motion when wheels do not slip at all. The parameters of the kinematic bicycle model, are presented in Table \ref{bicycleModel.tab}.

\begin{table}[h!]
\centering
\vspace*{0.15cm}

    \begin{tabular}{|c|l|l|}
      \hline
      \textbf{Variable} & \textbf{Characteristics} \\
      \hline
      $x$, $y$ & Coordinates in the map of the rear wheel \\
      $V$ & Velocity of the model at its rear wheel \\
      $L$ & Wheelbase of the vehicle \\
      $\delta$ & Steering angle \\
      $\psi$ & Yaw angle of the vehicle \\
      \hline
    \end{tabular}
    \caption{Characteristics of the kinematic bicycle model}
\label{bicycleModel.tab}
\end{table}

The state of the kinematic bicycle model is defined by $z=[x, y, \psi]^\top$, its input is defined by $u=[V, \delta]$. The state evolution is then defined by:
\begin{equation}\label{kinematicBicycle.eq}
    \dot z = f(z,u) = \begin{bmatrix}V\cos(\psi) & V\sin(\psi) & \frac{V \tan\delta}{L}\end{bmatrix}^\top
\end{equation}

We use a discretized version of this model to generate data and to build the EKF. Following the classical framework, we add process and measurement noises. Following Equation~(\ref{kinematicBicycle.eq}), the discrete time system is described by:
\begin{eqnarray}\label{discreteKinematicBicycle.eq}
 z_{k+1} &=& z_k + f(z_k, u_k).\Delta t + w_k\\
    m_k &=& z_k + v_k   
\end{eqnarray}
where $k$ is the time step, $\Delta t$ is the time interval, $m_k$ is the measurement, $w_k$ is the process noise and $v_k$ is the measurement noise. In order to stay close to the EKF assumptions, noises are white Gaussian with variance ${\sigma'}^2 = ({\sigma'}_x^2,{\sigma'}_y^2,{\sigma'}_\psi^2)$ for $w_k$ and $\sigma^2 = (\sigma_x^2,\sigma_y^2,\sigma_\psi^2)$ for $v_k$. Table~\ref{noise.tab} describes the noise parameters used.
The process noise is constant throughout all simulations. The measurement noise varies: the 3 standard deviations $(\sigma_x,\sigma_y,\sigma_\psi)$ are then all multiplied by a scaling factor $\alpha$. Our observer is trained using constant noise $\alpha=1$. Measurement noise levels are motivated by GPS/INS sensor properties presented in \cite{elkaim_comparison_2008}.

\begin{table}[h]
 \centering
 \begin{tabular}{|c|c|}
      \hline
      \textbf{Process}&\\
      \hline
      3$\sigma'_{x}$ & 0.2 m\\
        3$\sigma'_{y}$ & 0.2 m\\
        3$\sigma'_{\psi}$ & 3.4 mrad\\
        \hline
        \textbf{Measurement}&\\
        \hline
      3$\sigma_{x}$  & 1 m \\
        3$\sigma_{y}$  & 1 m\\
        3$\sigma_{\psi}$ & 17.4 mrad\\
        \hline
\end{tabular}
\caption{Noise parameters values. The measurement noise corresponds to $\alpha=1$.\label{noise.tab}}
\end{table}

\section{The Learning Based Observer}\label{3.sec}
We present here the core of our method to build our observer. First, we concentrate on the data generation, that is a particularly important step: we need to generate 3 sets of data: for training, where the main point is to get diversity; for validation, in order to stop the training and avoid over-fitting; and finally for testing. We use different data generation algorithms to minimize the risk of bias. Second, we introduce several architectures which observing performance will be evaluated.

In what follows, the kinematic bicycle model is assumed to be fully observable, with noisy measurements. The learning-based observer will be used to observe its actual states.
The data generation algorithms are presented in Section \ref{dataGen.ssec}, and the observers to be trained in Section \ref{arch.ssec}.

\subsection{Data Generation Algorithms}\label{dataGen.ssec}
We focus on generating a fairly distributed training data set over the behaviors of the vehicle; a validation data set that assesses the performance of the model while training; and a testing data set to test the learning-based observer after training it. The three generation algorithms are different to confirm the performance of the observer on different data sets. In what follows, we present the training data generation algorithm in Section \ref{trData.ssec}, the validation and testing data generation algorithms in Section \ref{valData.ssec}. 

\subsubsection{Training Data Generation}\label{trData.ssec}
Our training set should depict most of the behaviors of the vehicle represented in our case by the kinematic bicycle model. This will be represented by a fair distribution of the accelerations on a predefined friction circle. The friction circle represents an envelope for the
possible accelerations of a vehicle; and the position of an acceleration set of a vehicle in the friction circle determines the harshness of the effected behavior. The used friction circle has a radius of $0.5g$ representative of the domain in which the kinematic bicycle model can give a good representation of the model of the vehicle \cite{polack_kinematic_2017}. The distribution of the heading angles and the velocities will be taken into consideration as well.

A discrete simulator that implements the kinematic bicycle model is used based on Equation (\ref{discreteKinematicBicycle.eq}), and the goal is to determine the inputs that should be applied at each time step to generate the desired diverse data. Note that the defined process noise is implemented, low measurement noise is added after generating the data as we are dealing with an open loop controller. The kinematic bicycle model equations are differentiated to link the longitudinal and lateral accelerations to the inputs that should be applied to the vehicle. The differentiation of Equation (\ref{kinematicBicycle.eq}) in discrete time leads to: 
\begin{eqnarray}
    a_{x,k} = \frac{V_{k+1}-V_{k}}{\Delta t} \cos\psi_k - \frac{V_{k+1}\tan\delta_{k+1}}{L}V_k\sin\psi_k \\
    a_{y,k} = \frac{V_{k+1}-V_{k}}{\Delta t} \sin\psi_k +  \frac{V_{k+1}\tan\delta_{k+1}}{L}V_k\cos\psi_k
\end{eqnarray}
Where $a_{x,k}$, $a_{y,k}$ are the longitudinal and lateral accelerations at time step $t=k$, the remaining parameters being defined before, with $\Delta t = 0.02~s$. Having given longitudinal and lateral accelerations, the velocity and steering angle to be applied in the next time steps can be solved from the two above equations.

\begin{figure}[ht]
\begin{center}
\includegraphics[width=.8\columnwidth]{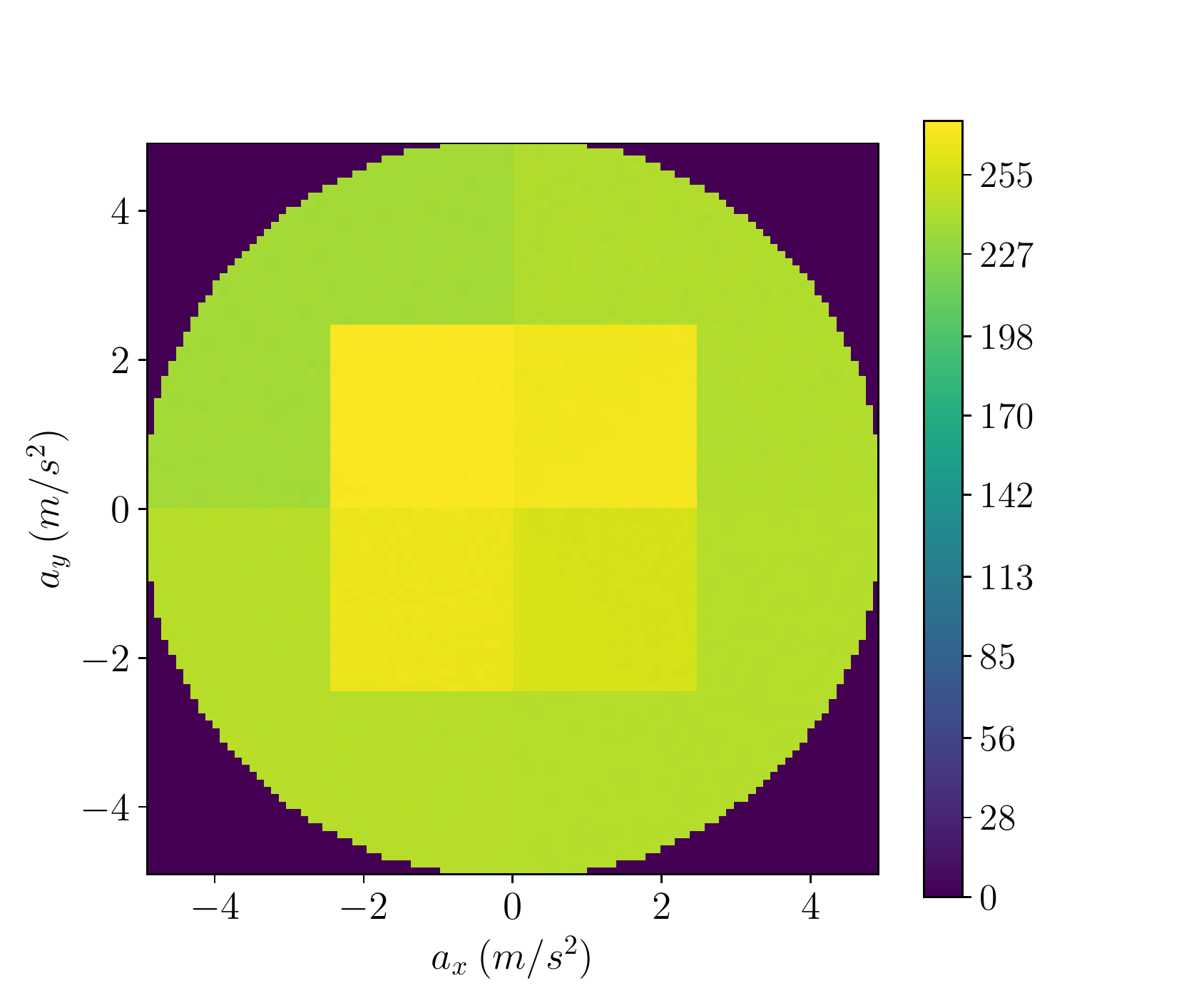}
\caption{The friction circle filled with the training data generation algorithm. The figure shows a fair distribution of the data over the circle: not even, but diverse enough. The color represents the density of samples.}
\label{frictionCircle.fig}
\end{center}
\end{figure}

The friction circle to be filled is represented by a 2D grid with a preassigned radius. Inputs to the model are chosen such that the resulting accelerations fill the less dense parts of the grid. One thousand 40-second trajectories are generated, each starting with a vehicle state $x=0$, $y=0$ and a randomly generated initial heading in the interval $\psi \in [-\pi, \pi[$.
Running the algorithm has generated a training data set made of 2 million samples and the distribution on the friction circle shown in Figure \ref{frictionCircle.fig}. Measurement noise in the training data set has a noise level of $\alpha=1$. We create the validation and testing data sets next.


\subsubsection{Validation and Testing Data Generation}\label{valData.ssec}
Validation data is used to give an unbiased evaluation of the network's performance during the training process; testing data is used to evaluate the performance of the network after training it. Both show how the learning-based observer is performing on data different than the training data. In our case, we use clothoid functions to create both sets. They result in smooth paths (C2 continuous). After generating the path to be followed, the pure-pursuit lateral controller \cite{coulter_implementation_1992} is used to follow it with a continuous velocity vector that changes according to random accelerations. The measurement noise in the validation data set is parametrized by a noise level of $\alpha=1$ while it varies with $\alpha \in [0;6]$ for the testing data set. For the validation data, we create a sinusoidal shape (using clothoids) alternating between $y = 5 \ m$ and $-5 \ m$ over an $x$ interval of $ 200 \ m$. The validation path is made of 995 data points. Its sole usage is to help define when to stop training, to avoid over-fitting. For the testing data, we create 15 testing trajectories representing challenging vehicle maneuvers to be evaluated, resulting in 9,829 data points. The testing trajectories are produced with variations of measurement noise levels, with noise scaling factor $\alpha$ varying between 0 and 6 with increments of 0.25: a total of 245,725 data points.

\subsection{The Observer Architecture}\label{arch.ssec}
After generating the needed data sets, the architecture of the learning based observer is to be defined. The learning based observer should take as input the state measurements $x_{m}^k$, $y_{m}^k$, $\psi_{m}^k$ and the inputs to the system $V^k$, $\delta^k$ for $k=t-n$, $n=0..N$, $t$ being the current time, $n$ being a time step, $N$ being a defined number of previous time steps. The output is the actual state of the kinematic bicycle model at $k=t$ (the generated data before noise addition).

We consider two main architectures: A Convolution Neural Network (CNN) and a Long Short-Term Memory (LSTM). For each of the networks we consider input window sizes of $N=20$, $N=40$, $N=60$ and $N=80$ time steps, a total of 8 learned observers which will be compared based on metrics defined later on. Details of the architectures are presented next.  

\subsubsection{CNN Architecture}
\begin{figure}[h]
\centering
\includegraphics[width=.9\columnwidth]{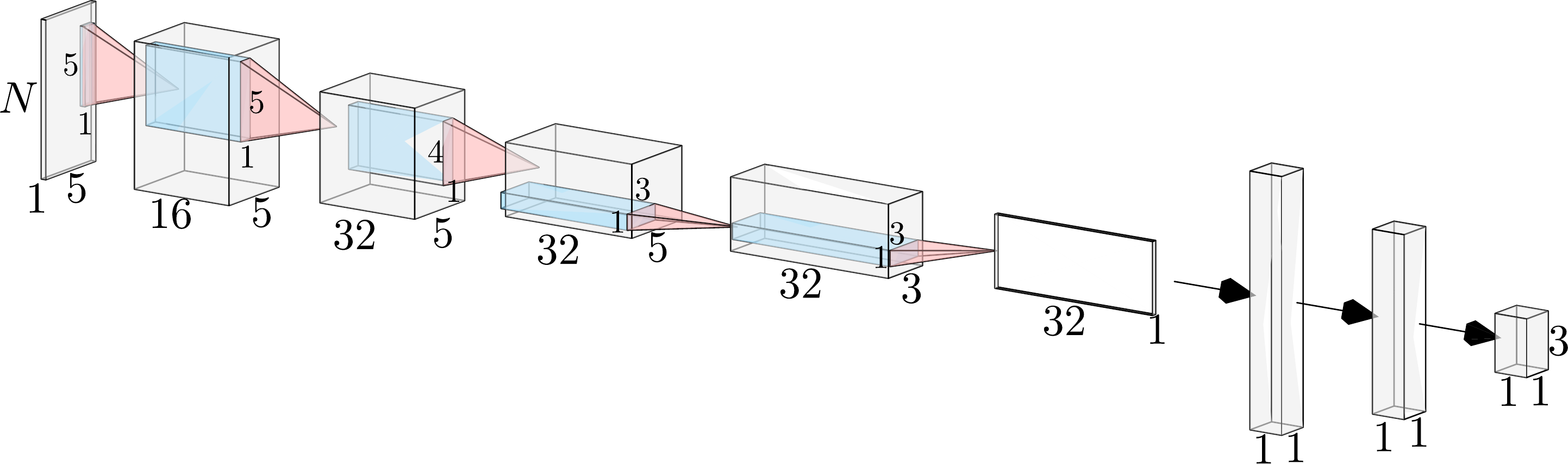}
\caption{CNN Architecture}
\label{arch1.fig}
\end{figure}

CNNs are neural networks that involve a series of convolution and pooling layers. They process data that has a grid-like topology, like the measurements and inputs for different time steps considered in this paper. Figure \ref{arch1.fig} shows the used architecture consisting of a CNN module followed by fully connected layers. The CNN module is described by the following:
\begin{subequations}
\begin{eqnarray}
    h^0 =& X &\\
    h^{(k)} =& \sigma^{(k)}(\pi^{(k)}(W^{(k)}*h^{(k-1)}+b^{(k)}))~~k=1..L
\end{eqnarray}
\end{subequations}
where $X$ is the input to the CNN module, $h^{(k)}$ is the output of layer $k$, $h^{(L)}$ is the output of the CNN module, $L$ is the number of layers, $\sigma$ is the used activation function, a sigmoid function in our case, $\pi$ is the pooling function and $W$ and $b$ are the corresponding weights and biases. The used fully connected layers are described by the following:  
\begin{subequations}\label{mlp.eq}
\begin{eqnarray}
    h^0 =& X &\\
    h^{(k)} =& \sigma^{(k)}(W^{(k)\top}h^{(k-1)}+b^{(k)}) & k=1..L'
\end{eqnarray}
\end{subequations}
where $X$ is the input vector, which is the output of the last CNN layer in our case, $h^{(k)}$ is the output of layer $k$, $h^{L'}$ is the output of the network, $L'$ is the number of layers, $\sigma$ is the used activation function, a sigmoid function in our case and $W$ and $b$ are the corresponding weights and biases.
The used architecture first extracts information from the different time steps involved so the network entails at first convolutions and pooling to reduce the length of the input while keeping the same number of features, then convolving to extract information from the different features. The network is made of: two convolution layers with filters of 5x1, followed by a max pooling layer of 4x1, and then two convolution layers with a filter of size 1x3; then 2 fully connected layers. Inputs are processed first separately, then merged. 

\subsubsection{LSTM Architecture}
\begin{figure}[h]
\centering
\includegraphics[width=.6\columnwidth]{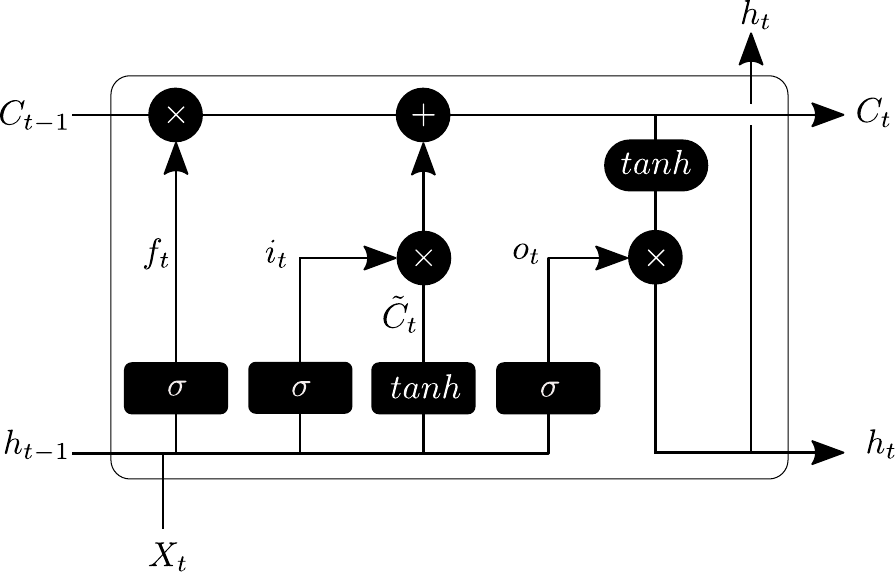}
\caption{LSTM Cell}
\label{lstmcell.fig}
\end{figure}

\begin{figure}[h]
\centering
\vspace*{0.2cm}
\includegraphics[height= .1\columnwidth, width=\columnwidth]{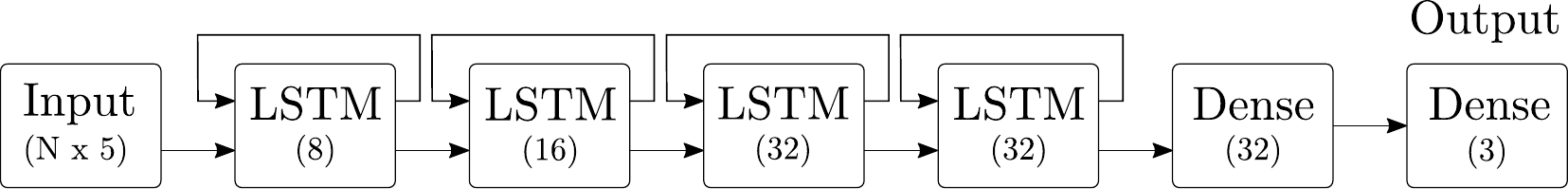}
\caption{LSTM Architecture}
\label{arch2.fig}
\end{figure}

LSTMs are a type of recurrent neural networks (RNNs). They are suited for time series, like the measurements and inputs for different time steps considered in this paper. A representation of a single LSTM cell is shown in Figure \ref{lstmcell.fig} and is described by the following equations:
\begin{subequations}
\begin{eqnarray}
    i_t &=& \sigma(W_i[h_{t-1},X_t]+b_i)\\
    f_t &=& \sigma(W_f[h_{t-1},X_t]+b_f)\\
    o_t &=& \sigma(W_o[h_{t-1}, X_t]+b_o)\\
    \Tilde{C}_t &=& \tanh(W_c[h_{t-1},X_t]+b_c)\\
    C_t &=& f_t \odot C_{t-1}+i_t \odot \Tilde{C}_t\\
    h_t &=& o_t \odot \tanh(C_t)
\end{eqnarray}
\end{subequations}
where $X_t$ is the input vector, $i_t$ is the input gate, $f_t$ is the forget gate, $o_t$ is the output gate, $C$ is the cell state, $h_{t-1}$ is the previous hidden state vector and $h_t$ is the output vector; $W_i$, $W_f$ $W_o$, $W_c$ are the weights and $b_i$, $b_f$ $b_o$, $b_c$ are the biases. The used architecture (shown in Figure \ref{arch2.fig}) uses four consecutive LSTM layers involving 8, 16, 32 and 32  LSTM neurons followed by 2 fully connected (dense) layers. The fully connected layers follow Equations (\ref{mlp.eq}) with $X$ being the output of the last LSTM layer. 

The loss function used is the mean squared error. The Xavier initialization is used to set the initial weights for the network. Hyperparameter optimization is done using grid search for $N=20$ for both architectures and then applied to $N=20$, $N=40$, $N=60$ and $N=80$. 

\section{Results and analysis}\label{4.sec}
After generating the required data sets and training the defined networks, the performance of the learned observers should be compared to the performance of the EKF. The testing set is made of data never seen by the trained networks in terms of the generated trajectories and the noise levels. The procedure will start by comparing the four observers of each architecture together; then, the best performing CNN observer and the best performing LSTM observer will be both compared to the EKF. In what follows the used metric will be detailed, followed by the comparison of the CNNs, the comparison of the LSTMs, and the comparison of the observers with the EKF. 

\subsection{Metric}
To compare the performance of the different observers a clear metric should be defined. Having a 3-dimensional error (1 dimension for each state variable $x$, $y$ and $\psi$) the error will be reduced to a scalar. For this purpose, we introduce the normalized root mean square error (NRMSE). It is normalized to 1 for the reference, i.e. for the root mean squared error (RMSE) of the EKF at low noise ($\alpha=1$). It allows to describe the performance of an observer with a single error. The introduced metric gives similar weight to the three variables. The normalized error metric is defined as:
\begin{equation}
    \text{NRMSE} = \sqrt{w_x E^2_x + w_y E^2_y + w_\psi E^2_\psi}
\end{equation}
$E^2_x$, $E^2_y$, $E^2_\psi$ being the mean square error for each of the three variables and $w_x$, $w_y$ and $w_\psi$ being the weights given to each variable based on the reference EKF low noise case, such that $w_x = \frac{1}{3E^2_{ref, x}}$, $w_y = \frac{1}{3E^2_{ref, y}}$, $w_\psi = \frac{1}{3E^2_{ref, \psi}}$, with: $E_{ref, x}=0.24~m$, $E_{ref, y}=0.23~m$ and $E_{ref, \psi}=4.1~mrad$.  Having the metric to be used, the comparison between the different observers is presented next.

\subsection{CNN observers comparison}
The defined metric is applied to the predictions of the four CNN observers on the previously defined testing data set. The plot in Figure \ref{CNN_NRMSE.fig} shows the evolution of the four observers with respect to noise increase. It can be seen that for all observers, the performance deteriorates with the increase of the noise level. The $N=20$ and $N=80$ observers have close performance. The $N=60$ observer shows lower errors for all noise levels and will be used for the comparison with the EKF.  

\subsection{LSTM observers comparison}
The NRMSE metric is as well applied to the predictions of the four LSTM observers on the  previously defined testing data set. The plot in Figure \ref{LSTM_NRMSE.fig} shows the evolution of the four observers with respect to noise increase. LSTM observers show more robustness to noise than CNN observers but have higher errors. The $N=80$ and $N=20$ observers show a close performance especially for higher noise levels. The $N=60$ observer shows the highest errors. The $N=80$ observer has the lowest errors and will be used for the comparison with the EKF.  


\subsection{Comparison with the EKF}
\begin{figure*}[t!]
\centering
\subfloat[NRMSE plot for the four CNN based observers. The $N=60$ observer has the lowest errors.]{\includegraphics[width=.66\columnwidth ]{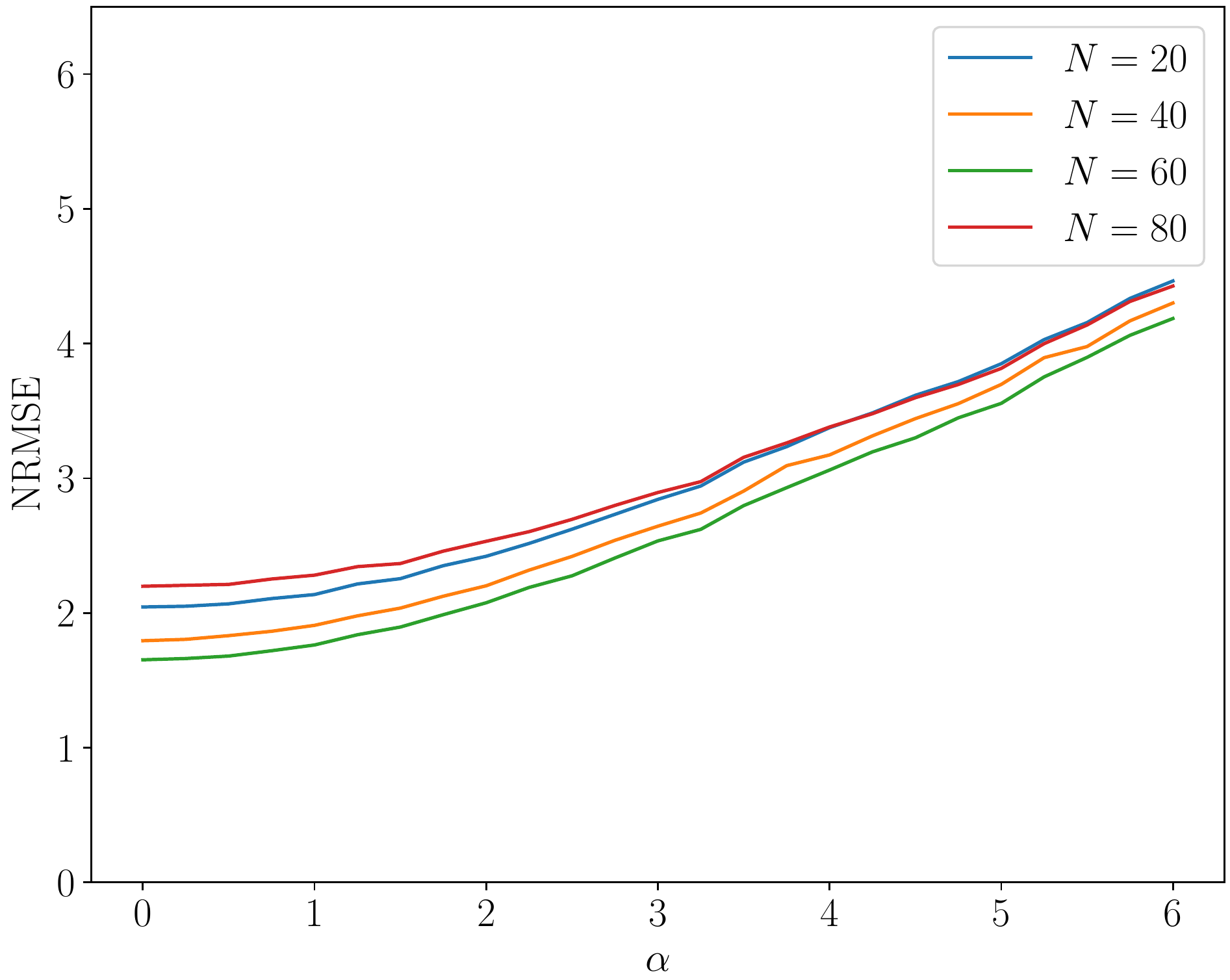}\label{CNN_NRMSE.fig}}
\hfill
\centering
\subfloat[NRMSE plot for the four LSTM based observers. The $N=80$ observer has the lowest errors. ]{\includegraphics[width=.66\columnwidth ]{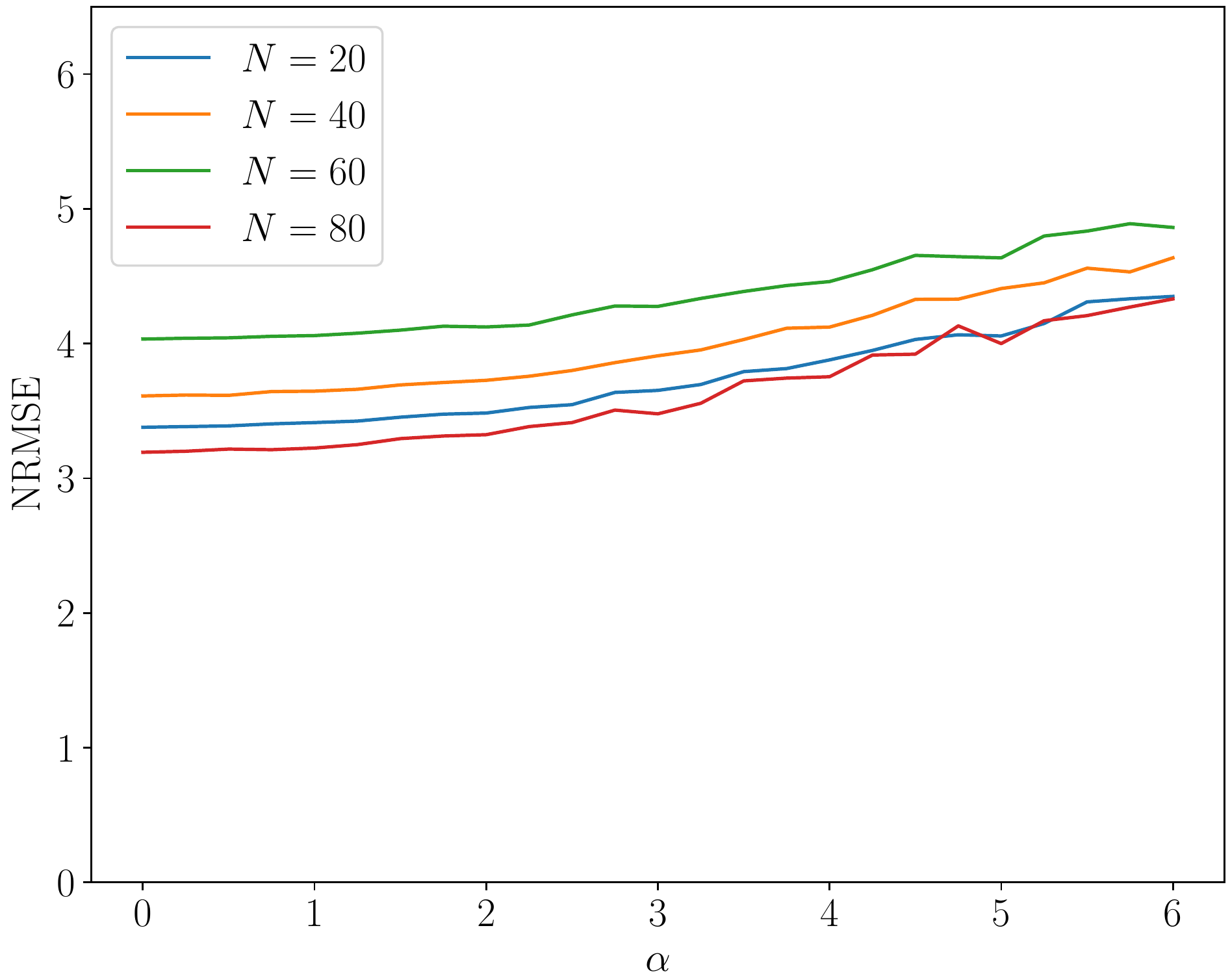}\label{LSTM_NRMSE.fig}}
\hfill
\centering
\subfloat[NRMSE comparison between the $N=80$ LSTM, $N=60$ CNN, and EKF observers. The CNN outperforms the LSTM and surpasses the EKF with noise increase.]{\includegraphics[width=.66\columnwidth]{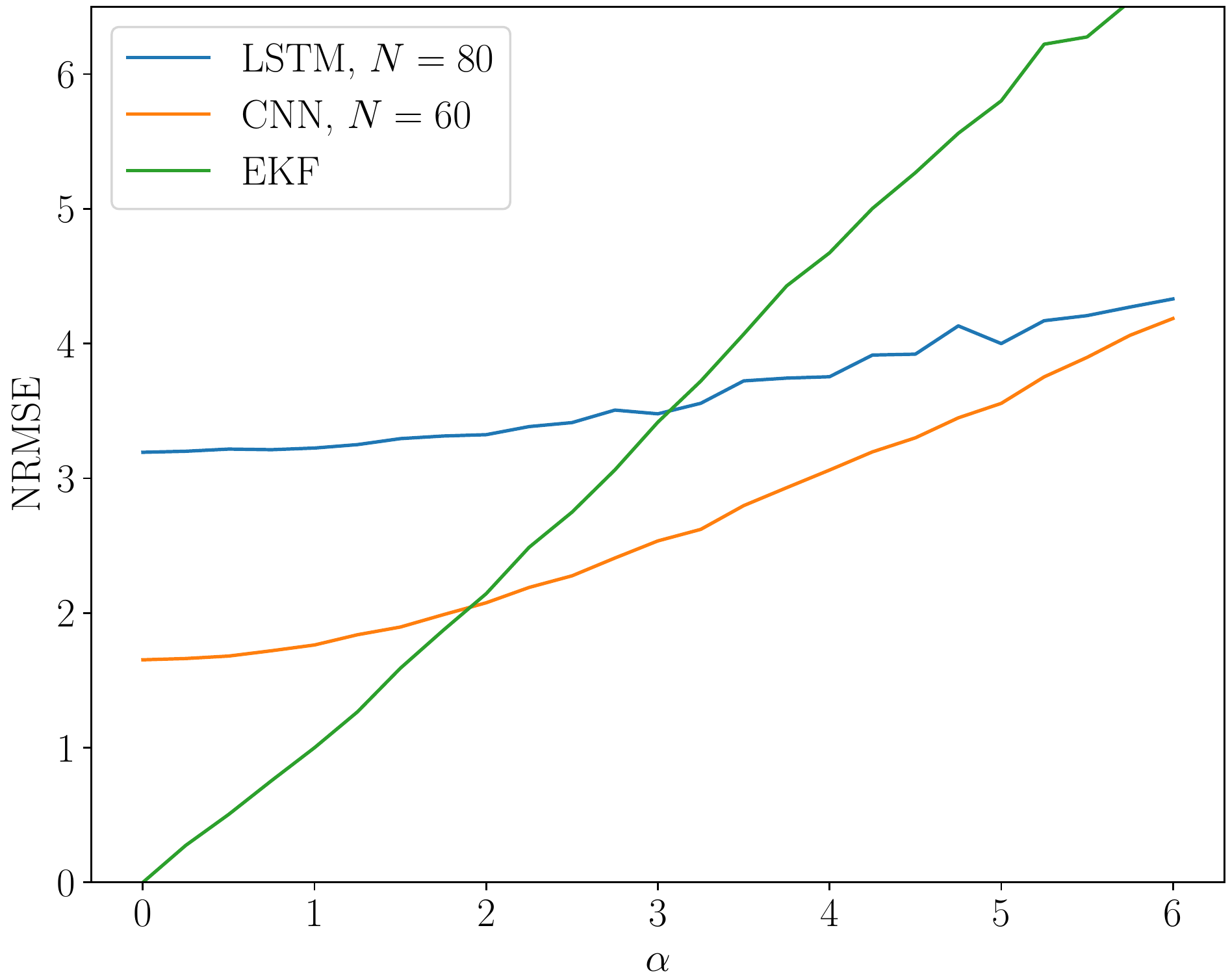}\label{ALL_NRMSE.fig}}
\hfill
\caption{NRMSE comparison plots between different observers.}
\end{figure*}

\begin{figure}
\centering
\includegraphics[width=.8\columnwidth]{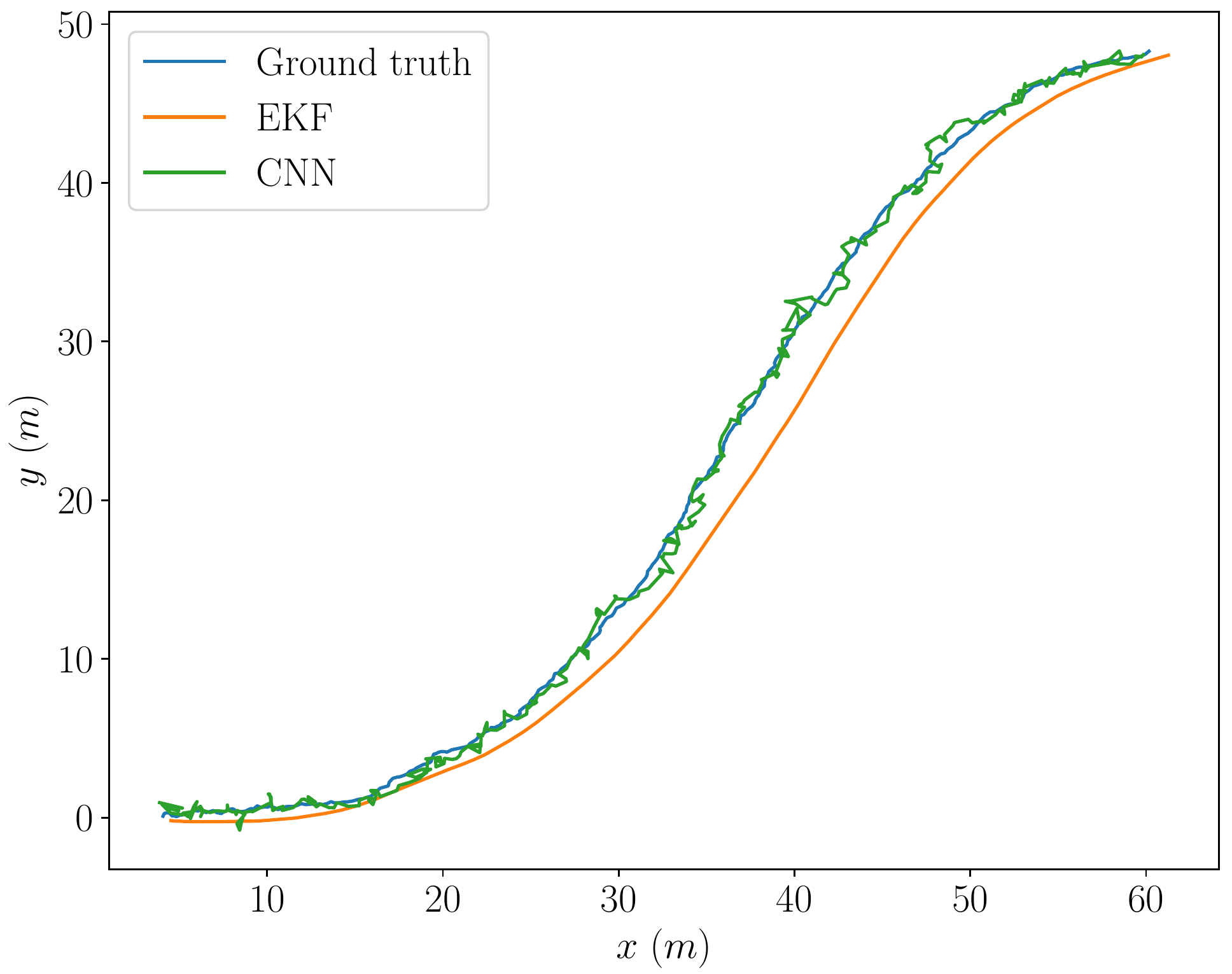}
\caption{Comparison between the $(x,y)$ estimations of  the CNN observer and the EKF for a sample testing trajectory with $\alpha=3$. The estimates of the CNN observer are overall closer to the ground truth than the estimates of the EKF.}
\label{testingPath.fig}
\end{figure}

Having compared the different models with different input window shapes, the best performing observers can be compared with the EKF. Tables \ref{rmse1.tab} and \ref{rmse2.tab} present the RMSE for the three predicted variables for each of the observers for $\alpha=1$ and $\alpha=6$ respectively. The tables show that the EKF clearly outperforms the learned observers for $\alpha=1$ while it is outperformed by both learned observers for $\alpha=6$. It is remarked as well that for the $x$ and $y$ predictions the CNN clearly outperforms the LSTM while the performance is close for the $\psi$ predictions. The NRMSE metric defined above is used to compare the three observers. The plot in Figure \ref{ALL_NRMSE.fig} shows the performance of the three observers in terms of the noise scaling factor $\alpha$. It can be seen that for lower noise the EKF performs better than the two learned observers. Both learned observers show their best performance in low noise domains, which is logical as they have been trained on low noise data. Though, their performance in these domains cannot beat that of the EKF. But, the performance of the EKF deteriorates gradually (almost linearly) with the increase of the noise scaling factor, showing low robustness to noise increase. On the other hand, learned observers show higher robustness to noise increase. The CNN based observer shows lower NRMSE scores than the LSTM one for all levels of noise. It is estimated that the CNN outperforms the EKF at $\alpha\simeq 2$ and that the LSTM outperforms the EKF at $\alpha\simeq 3$. A sample of the performance difference between $(x,y)$ estimations of the CNN and the EKF for $\alpha = 3$ is shown in Figure \ref{testingPath.fig}: the CNN estimations are more accurate than the EKF estimations. The presented results and our analysis confirm a rather logical behavior: for low measurement noise, meaning the measurements are reliable, model-based observers are able to overcome the noise more efficiently than our learned-based observers. However, the learned observers deal better with higher levels of measurement noise, meaning when measurements are less reliable. 

In brief, the proposed CNN architecture is able to outperform the LSTM architecture for all the considered levels of measurement noise, and to outperform the EKF at a specified measurement noise domain.

\begin{table}[ht]
 \centering
  \begin{tabular}{|c|c|c|c|}
    \hline
  \textbf{Variable} & \textbf{EKF} & \textbf{CNN} & \textbf{LSTM} \\
      \hline
      $x$ (m) & 0.24 & 0.28 & 0.65 \\
        $y$ (m) & 0.23 & 0.23 & 0.90 \\
        ${\psi}$ (mrad) & 4.1 & 11 & 13\\
        \hline
        \end{tabular}
  \caption{RMSE comparison of the observers on low noise data ($\alpha=1$). The EKF shows the lowest errors. The CNN performs better than the LSTM. \label{rmse1.tab}}
\end{table}

\begin{table}[ht]
 \centering
  \begin{tabular}{|c|c|c|c|}
    \hline
  \textbf{Variable} & \textbf{EKF} & \textbf{CNN} & \textbf{LSTM} \\
      \hline
      $x$ (m) & 1.56 & 0.91 & 0.98\\
        $y$ (m) & 1.88 & 0.89 & 1.10 \\
        ${\psi}$ (mrad) & 26 & 20 & 19\\
        \hline
        \end{tabular}
  \caption{RMSE comparison of the observers on high noise data ($\alpha=6$). The EKF shows the highest errors. The CNN performs better than the LSTM except for $\psi$ prediction where the performance is close.\label{rmse2.tab}}
\end{table}

\section{Conclusion}\label{5.sec}
In this paper, we presented a method to build and train a learning-based observer. A kinematic bicycle model was used to generate its training, validation and testing datasets. We focused on generating an unbiased dataset in terms of accelerations to be representative of the full behavior of the model, while keeping the model valid. Two networks were considered: CNN and LSTM, with different considerations to the input window shape. The best performing observers were then compared to an EKF observer based also on a kinematic bicycle model for the prediction step. The results showed that our approach was able to outperform the EKF when the measurement noise increases which confirms our point: learning-based observers can outperform classical ones even for simple models. The CNN architecture outperformed the LSTM one. The CNN architecture can be adapted to many systems even with more inputs than in our study.

A point for future analysis is the inability of the learned observers to take benefit of low measurement noises. It may be due to a lack of diversity in the training, regarding noise levels: the network was trained with a constant noise level. This shows the friction circle probably does not describe enough the needed diversity for training.

Future work will focus on applying learning-based observers to real vehicles and considering more complex vehicle models. These models are challenging for classical observers, because the more details it catches, the more parameters it gets, leading to increased model errors. Also, mathematical derivation of observers becomes tricky, due to many non-linearities, while our CNN structure can scale easily. Creating diverse enough training data becomes more challenging.

\let\oldbibliography\thebibliography
\renewcommand{\thebibliography}[1]{%
  \oldbibliography{#1}%
  \setlength{\itemsep}{0pt}%
}

\bibliographystyle{ieeetr}
\bibliography{citations}

\end{document}